\pdfoutput=1

\documentclass[11pt]{article}

\usepackage[]{acl}

\usepackage{times}
\usepackage{latexsym}

\usepackage[T1]{fontenc}

\usepackage[utf8]{inputenc}

\usepackage[subtle]{savetrees}
\usepackage{microtype}
\makeatletter
\MT@redefine@patch{eqnum}{%
  \MT@patch@patch\tagform@{(}{\leftprotrusion{(}}%
  \MT@patch@patch\tagform@{)}{\rightprotrusion{)}}%
  \MT@ifdefined@n@TF{eqref }
    {\MT@exp@cs\MT@patch@patch{eqref }}{\MT@patch@patch\eqref}
       {\tagform@}{\@nameuse{MT@patch@saved@\string\tagform@}}%
}{}
\makeatother

\usepackage{url}
\usepackage{times}
\usepackage{latexsym}

\usepackage{multicol}
\usepackage{multirow}
\usepackage{graphicx}
\usepackage{verbatim}
\usepackage{makecell}
\usepackage{nicefrac, xfrac}
\usepackage{braket}
\usepackage{hhline}
\usepackage{amsmath,bm,amssymb}
\usepackage[linesnumbered,ruled]{algorithm2e}
\usepackage{xcolor}
\definecolor{Green}{rgb}{0.0, 0.5, 0.0}
\definecolor{Amethyst}{rgb}{0.6, 0.4, 0.8}
\title{Generating EDU Extracts for Plan-Guided Summary Re-Ranking }

\author{
  Griffin Adams$^{\spadesuit,\clubsuit}$ \\ griffin.adams@columbia.edu \And
  Alexander R. Fabbri$^{\diamondsuit}$ \\ afabbri@salesforce.com \\ \And Faisal Ladhak $^{\spadesuit}$ \\ faisal@cs.columbia.edu \\ \AND
  Kathleen McKeown $^{\spadesuit}$ \\ kathy@cs.columbia.edu \\\And No\'emie Elhadad$^{\spadesuit,\clubsuit}$ \\ noemie.elhadad@columbia.edu \AND
 Salesforce Research$^{\diamondsuit}$ \quad  Columbia University: Computer Science$^{\spadesuit}$, Biomedical Informatics$^{\clubsuit}$
}

\begin{document}
\maketitle
\begin{abstract}

Two-step approaches, in which summary candidates are generated-then-reranked to return a single summary, can improve ROUGE scores over the standard single-step approach. Yet, standard decoding methods (i.e., beam search, nucleus sampling, and diverse beam search) produce candidates with redundant, and often low quality, content. In this paper, we design a novel method to generate candidates for re-ranking that addresses these issues.  We ground each candidate abstract on its own unique content plan and generate distinct plan-guided abstracts using a model's top beam. More concretely, a standard language model (a BART LM) auto-regressively generates elemental discourse unit (EDU) content plans with an extractive copy mechanism. The top $K$ beams from the content plan generator are then used to guide a separate LM, which produces a single abstractive candidate for each distinct plan. We apply an existing re-ranker (BRIO) to abstractive candidates generated from our method, as well as baseline decoding methods. We show large relevance improvements over previously published methods on widely used single document news article corpora, with ROUGE-2 F1 gains of $\bm{0.88}$, $\bm{2.01}$, and $\bm{0.38}$ on CNN / Dailymail, NYT, and Xsum, respectively. A human evaluation on CNN / DM validates these results. Similarly, on 1k samples from CNN / DM, we show that prompting GPT-3 to follow EDU plans outperforms sampling-based methods by $\bm{1.05}$ ROUGE-2 F1 points. Code to generate and realize plans is available at \url{https://github.com/griff4692/edu-sum}.

\end{abstract}

\section{Introduction}

Generating diverse abstracts and then re-ranking can lead to large performance gains (in ROUGE) \citep{liu-etal-2022-brio, ravaut-etal-2022-summareranker} over the standard approach of generating a single summary. Typically, diversity is controlled for at the \emph{token}-level by modifying beam search to introduce sampling (top-K \citep{fan-etal-2018-hierarchical}, nucleus \citep{holtzman2019curious}) or directly penalize repetition \citep{vijayakumar2016diverse}.


Yet, there is a tradeoff, as these methods tend to achieve diversity at the expense of quality~\citep{holtzman2019curious}. To avoid content de-generation while still achieving diversity\footnote{While highly important, in this work, we focus on content selection, not on the faithfulness of model-generated summaries.}, diversity can be introduced during a planning stage, as in \citet{narayan-etal-2022-well}, who generate entity chain plans with diverse beam search before realizing a summary with regular beam search.

In this paper, we also explore achieving diverse summaries through diverse plans, yet we focus on grounded extractive plans, which promote diversity by encouraging a model to focus on specific, unique parts of the source text. We define a content plan as a set of non-overlapping text spans from the source document. Specifically, we choose elemental discourse units (EDUs) as the appropriate granularity for content planning \citep{mann1988rhetorical}. EDUs represent sub-sentential independent clauses and allow for more fine-grained control than sentence-level extraction. EDUs are more self-contained and less fragmented than other potential sub-sentence content units, e.g. entities or noun phrases. Extractive EDUs are contiguous and are atomic, whereas entities do not cover all content and can appear in multiple contexts.


\begin{figure}[t]
\centering
\includegraphics[width=\linewidth]{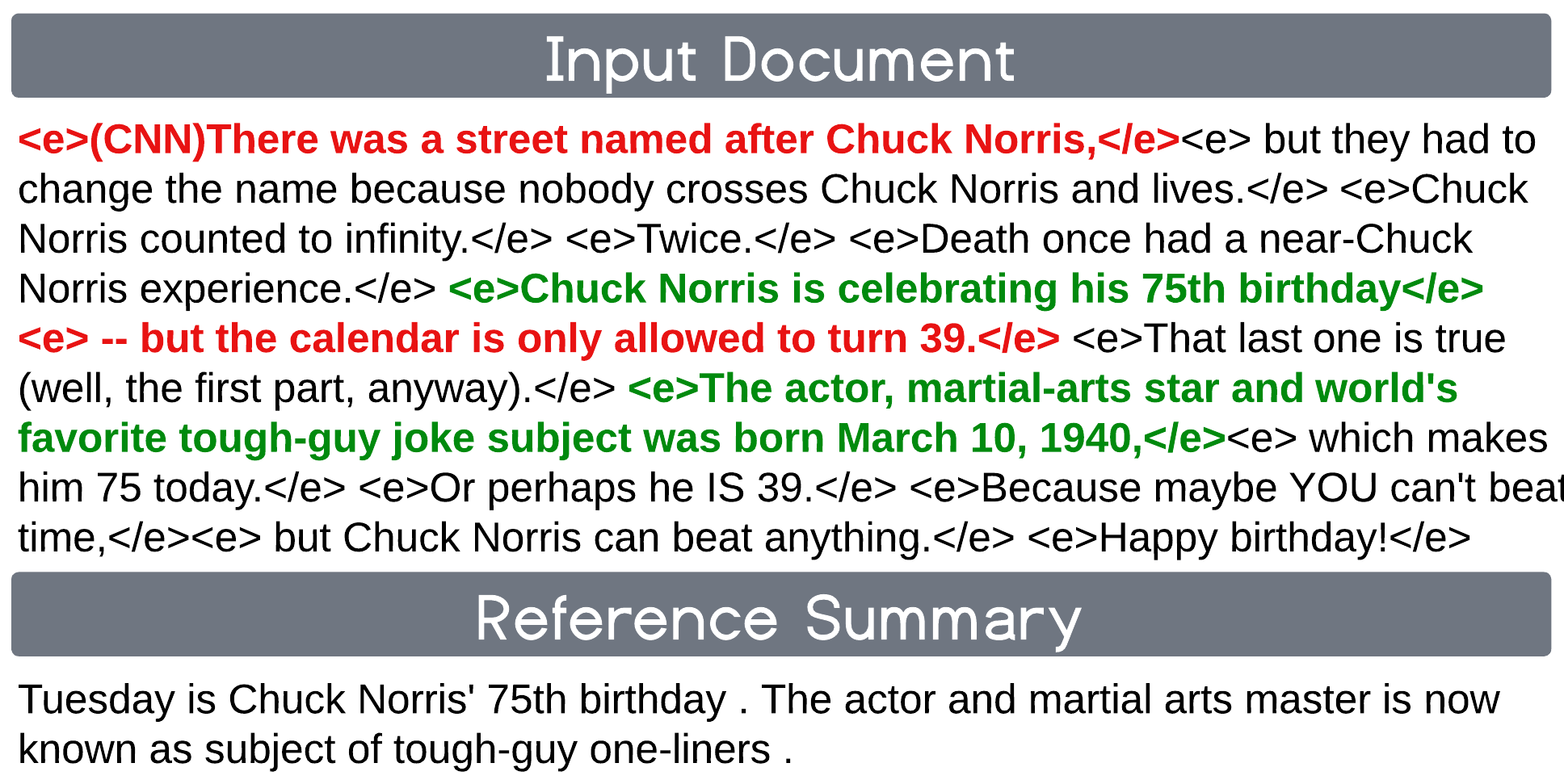}
\caption{ EDU Plan-Guided Abstraction (PGA). \textcolor{Green}{EDU spans} form the oracle content plan, while \textcolor{red}{EDU spans} form a random distractor plan. A model is trained to generate the reference only when given the oracle plan, not the random one. EDU-level plans afford more fine-grained control than sentence-level as irrelevant content is cut out: ``but the calendar is only allowed to turn 39''. }
\label{fig:plan-example}
\vskip -0.1in
\end{figure}

At a high-level, we employ two encoder-decoder models. Given a document, the first model generates $K$ unique content plans with beam search. Then, each content plan is used as a guide to a second model, which realizes an abstract given the plan and the document. Specifically, a BART-based \citep{lewis-etal-2020-bart} hierarchical encoder-decoder learns to generate extracts from left-to-right by copying EDUs until a special end of extract token is copied. These extractive plans are used to decorate the input document and serve as a guide for the Plan-Guided Abstractor (PGA). The top $K$ beams are returned from the content planner, while only the top beam is returned for plan realization to avoid de-generation. An example of the training procedure from the CNN/DailyMail news dataset is shown in Figure \ref{fig:plan-example}.

We compare our PGA candidate generation method to other decoding baselines (beam search, diverse beam, search, and nucleus sampling) at both the candidate level (across beams), as well as after applying a re-ranker (BRIO \citep{liu-etal-2022-brio}) to obtain a single, re-ranked summary.  We also benchmark the performance of re-ranked summaries from our PGA method against publicly reported results from other summary re-ranking papers. We note consistently higher ROUGE and BERTScores against both our internal baselines and public benchmarks, which we link to improved content selection across candidate beams. We also conduct a human evaluation and find that annotators assess top ranked summaries from PGA candidates as containing more relevant content than candidates produced by baseline decoding methods. By separately optimizing the plan and plan-guided abstracts, we can easily combine generated plans with a Large Language Model (LLM). In \S \ref{sec:gpt}, we prompt GPT-3.5 to generate diverse, \emph{focused} summaries and apply a re-ranker. We compare with a series of \emph{un-focused} prompts and find that ROUGE scores improve across the board. More generally, prompting with diverse plans, and then re-ranking, is a convenient alternative to RLHF alignment when using closed models.

Our primary contributions are: \textbf{(1).} We propose a novel two-stage model for generating high-quality, diverse candidate summaries for downstream re-ranking. Our plan generation approach adapts a pre-trained LM to perform span-level copying to produce EDU-level plans. \textbf{(2).} Our plan-guided abstraction model leads to large improvements in top-ranked summaries vis-a-vis previously published results ($\bm{0.88}$, $\bm{2.01}$, and $\bm{0.38}$ ROUGE-2 F1 percentage point gains on CNN/DM, NYT, and Xsum, respectively), and outperforms on summary relevance according to human evaluation. \textbf{(3)} We perform extensive analysis of candidate generation methods, according to the diversity of derived content plans and factors, such as source length. \textbf{(4)} We show that we can improve the reference-based performance of few-shot LLMs by prompting for diverse summaries based on extractive EDU plans.

\section{Related Work}

\paragraph{Two-Step Summarization.}

Re-ranking candidate summaries can address the ``exposure bias'' problem \citep{ranzato2015sequence} from standard maximum likelihood teacher forcing by allowing an external model to coordinate system outputs with evaluation metrics. Re-ranking diverse candidates can lead to improved faithfulness \citep{zhao-etal-2020-reducing, chen-etal-2021-improving} or relevance (as measured by ROUGE) \citep{liu-liu-2021-simcls, ravaut-etal-2022-summareranker, liu-etal-2022-brio, zhao2022calibrating}. Ranking can also be incorporated into training by adding a contrastive loss to the standard MLE loss for a multi-task objective \citep{nan-etal-2021-improving, liu-etal-2022-brio}. This work is related to, yet distinct from, our work, as we focus on the impact of candidate generation methods on explicit re-ranking.

\paragraph{Diverse Decoding.}

Diverse candidates are typically generated by a pre-trained model by modifying standard beam search to introduce sampling (top-k \citep{fan-etal-2018-hierarchical} or a dynamic nucleus \citep{holtzman2019curious}) or penalizing repeated tokens across distinct beam groups \citep{diverse-beam}. While increasing diversity, these methods introduce a quality-diversity tradeoff \citep{ippolito-etal-2019-comparison}.


Our approach to generating diverse abstracts has similarities to Compositional Sampling, introduced by \citet{narayan-etal-2022-well}. They use diverse beam search to predict an entity chain--based on the authors' FROST model \citep{narayan-etal-2021-planning}, before continuing to decode with regular beam search. Sampling at the plan level encourages diversity without having to use degenerative token-level sampling. Our approach is different in that, rather than use entity chains, we explicitly control the content focus to specific sentence fragments (EDUs). The goal of their work is high quality diverse summaries, while the goal of our work is to leverage diversity to achieve a single high quality summary.

More concretely, we differentiate our approach along three dimensions. \textbf{(1) Uniqueness.} Composition Sampling uses diverse beam search (DBS) to construct an entity chain and a summary. DBS penalizes repetition across beam groups at the same position, which allows for nearly identical plans with shifted word order. FROST does not localize each entity, which may be problematic for documents with co-referent entities. Our approach performs beam search over discrete plans. As such, it enforces that each plan is unique and localized. \textbf{(2) Completeness.} Entities–a subset of noun phrases–do not cover all the information in a document. Our method considers contiguous spans with no gaps. \textbf{(3) Complementarity.} The top beam from the FROST model represents the highest joint likelihood of plan and summary. Given the length mismatch of summaries vs plans, the top beam may not return an optimal plan. Our EDU generator serves as a standalone planner, which makes it more easily integrated with an LLM, as we explore in \S \ref{sec:gpt}.


\paragraph{Extract-Then-Abstract}

 Methods that decouple content selection from surface realization have proven effective, especially for long-document corpora with high compression ratios \citep{pilault-etal-2020-extractive}. While typically a two-step, coarse-to-fine framework \citep{liu2018generating, zhang-etal-2022-summn}, end-to-end systems are possible by bridging the gap with latent extraction \citep{mao-etal-2022-dyle} or using reinforcement learning: optimizing ROUGE-based rewards with policy gradients \citep{chen-bansal-2018-fast} (Actor Critic), or multi-armed bandits \citep{song-etal-2022-improving} (Self-Critical).

For shorter tasks, two-step approaches have also proven effective \citep{mendes-etal-2019-jointly}.  Yet, given that input compression is less of a concern, extractive guidance can also be \emph{added} as an auxiliary input in a dual-encoder setup \citep{dou-etal-2021-gsum}. Guidance can either be provided as input (encoder-side \citep{he2020ctrlsum}) or generated as part of a decoder prompted content planning step \citep{narayan-etal-2021-planning}.

Our work is based on a two-step extract-then-abstract framework, yet the goal is very different. We use extraction, not just as a guide, but as a tool to control the diversity of downstream abstracts.

\begin{figure}[t]
\centering
\includegraphics[width=\linewidth]{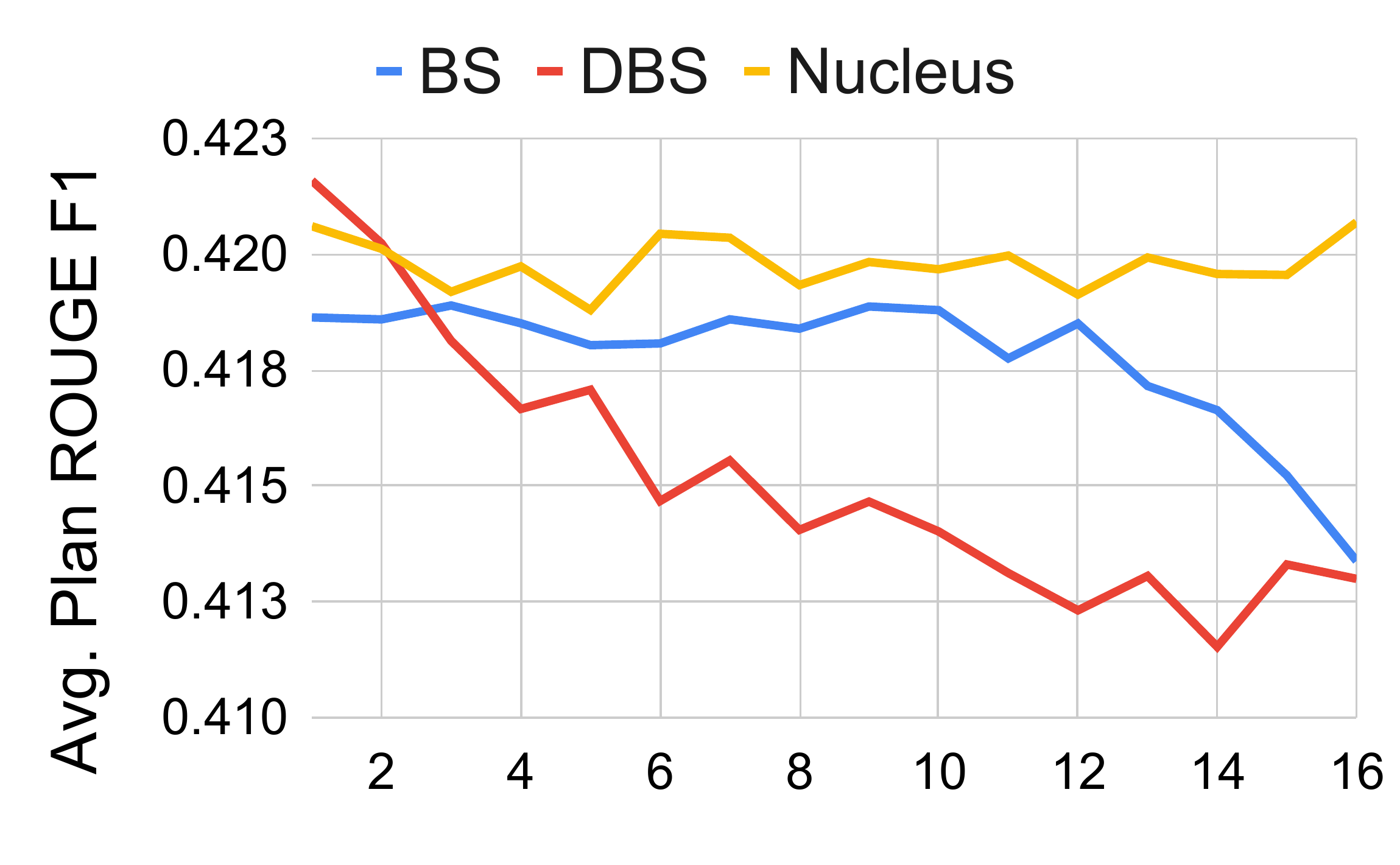}
\caption{ The average \textbf{Salience} of Derived Content Plans (DCPs) at different beams for BS (beam search), DBS (diverse beam search), and nucleus, or Top-P, sampling. Results shown are on the full CNN/DailyMail test set. }
\label{fig:salience}
\vskip -0.1in
\end{figure}

\begin{figure}[t]
\centering
\includegraphics[width=\linewidth]{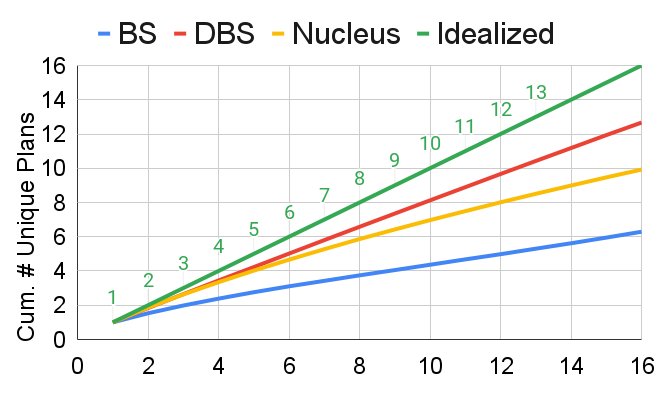}
\caption{ The \textbf{Uniqueness} score as a function of the beam size. Results shown are on the full CNN/DailyMail test set. }
\label{fig:uniqueness}
\vskip -0.1in
\end{figure}

\section{Motivation \& Analysis} \label{sec:motivation}
\paragraph{Elemental Discourse Units.} Prior work has shown that reference summary sentences usually combine information from multiple document sentences, while removing non-essential descriptive details \cite{lebanoff-etal-2019-scoring,liu-chen-2019-exploiting,li-etal-2020-composing}. As such, an ideal extractive plan would select only the relevant subsentential units to incorporate into the final summary. To achieve this, we rely on discourse level segmentation from Rhetorical Structure Theory \cite{mann1988rhetorical} to segment document sentences into Elementary Discourse Units (EDUs), which are contiguous spans of tokens representing independent clauses. EDUs are a good approximation \cite{li-etal-2016-role} of Summary Content Units (SCUs) written by human annotators for the Pyramid evaluation method \citep{nenkova2004evaluating}.

To extract EDUs, We use the neural parser \citep{liu2020multilingual, liu-etal-2021-dmrst}, fine-tuned from \texttt{xlm-roberta-base} \citep{conneau-etal-2020-unsupervised} on RST treebanks from 6 languages, to segment sentences into non-overlapping, contiguous EDU fragments. Their model merges short EDUs ($< 5$ tokens) to prevent fragmentation. As such, these EDU fragments are closer to proposition-level extraction than other possible units of extraction, e.g., entities.

\begin{table}[h]
\centering
\small
\begin{tabular}{l|cc|c}
\textbf{Text Unit} & \textbf{\# in Doc} & \textbf{\# in Oracle} & \textbf{Rouge-1 F1} \\ \hline
\textbf{Sentences} & 29.2 & 3.3 & 57.8 \\
\textbf{EDU} & 51.6 & 5.3 & 61.7 \\
\end{tabular}
\caption{ Comparing oracles formed from source sentences versus EDU spans on the CNN / Dailymail validation set. } \label{tab:edu-sent-comparison}
\vskip -0.1in
\end{table}

Table \ref{tab:edu-sent-comparison} displays statistics for EDU versus sentence segmentation. There are less than 2 EDUs per sentence ($\sfrac{51.6}{29.2}$) and less than 2 times as many EDUs in oracle extracts ($5.3$) as with sentences. Extractive oracles are computed the same way for both sentences and EDUs: by greedily selecting extractive units to maximize the average ROUGE-1 and ROUGE-2 F1 of partially built extracts against the reference summary, as in \citet{nallapati2017summarunner}. We compute the ROUGE-1 F1 overlap against the reference of oracles formed from EDUs versus sentences. EDUs outperform sentences ($61.7$ versus $57.8$), which confirms similar oracle analysis on CNN/DM from \citet{liu-chen-2019-exploiting}.


\paragraph{Content Selection Shortcomings of Existing Methods.} We first propose two simple preferred properties of candidate sets for re-ranking. The first is a \textbf{Salience Property}: all candidates should focus on relevant content. The rationale is trivial: a re-ranker will not always select the best candidate\footnote{In fact, \citet{liu-etal-2022-brio} note that even well-tuned re-rankers have a fairly low correlation with ROUGE scores.}, so it is important that, on average, candidates be relevant. The second is a \textbf{Uniqueness Property}: candidates should focus on different parts of the source. Without content diversity, there is limited upside to re-ranking over just taking the top beam. Because summaries are typically evaluated against a single reference, a tradeoff exists. High \textbf{Salience} favors candidates clustered around the reference, while \textbf{Uniqueness} favors exploration.

To quantify these properties, we introduce the notion of a \textbf{Derived Content Plan} (DCP). First, we align each summary to a set of extractive fragments from the source text (EDUs). We use a greedy approach, which maximizes the relative average ROUGE-1/ROUGE-2 F1 gain of adding each additional EDU from the source text to the plan.  This procedure is identical to the widely-used oracle sentence labeling defined by \citet{nallapati2017summarunner}, except that EDUs are extracted, not sentences. The unordered set of EDUs aligned to a summary form its \textbf{DCP}. Roughly speaking, DCPs map the content of each summary, which may exhibit some lexical variation, onto a shared space (the input document).

For this analysis, we then define \textbf{Salience} as the ROUGE-1 F1 overlap between a summary's DCP and the gold-standard reference. \textbf{Uniqueness}, on the hand, we define at the candidate set level. Specifically, it is the number of unique DCPs among a set of candidate summaries. Lower scores signal more content redundancy. Figure \ref{fig:salience} reveals a near monotonic decline in DCP \textbf{Salience} at each successive beam for beam search (BS) and diverse beam search (DBS). Nucleus sampling is constant given that each candidate is sampled independently. Figure \ref{fig:uniqueness} shows an \textbf{\textcolor{Green}{\texttt{Idealized}}} scenario in which $y=x$ and each candidate has a unique DCP. All baseline methods fall below the \textbf{\textcolor{Green}{\texttt{Idealized}}} line and exhibit DCP redundancy.

\begin{figure*}[t]
\centering
\includegraphics[width=\linewidth]{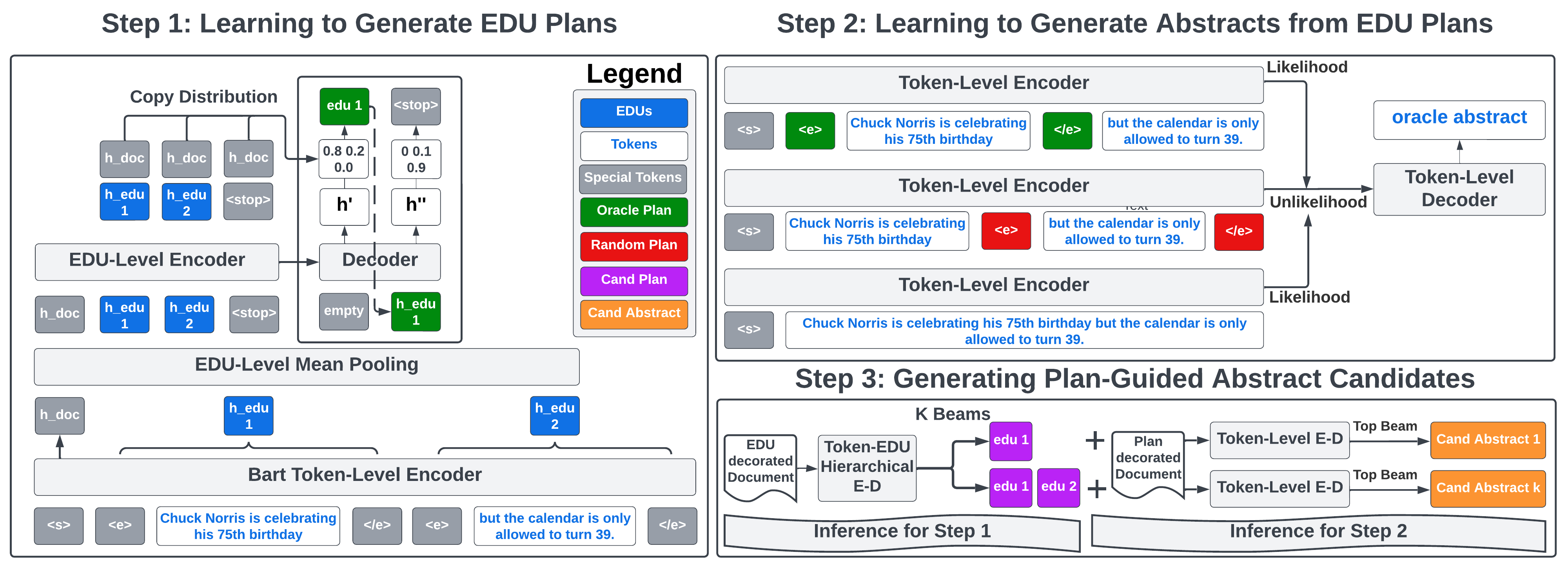}
\caption{ Plan-Guided Abstraction (PGA). In the first step, a token-level encoder processes a document decorated with special EDU boundary markers. EDU-level hidden states are formed with mean-pooling and serve as the inputs to a shallow EDU-level Encoder-Decoder, which learns to auto-regressively copy oracle EDU plans. In the second stage, a Plan-Guided Abstractor learns to generate abstractive reference summaries from inputs decorated with EDU boundary markers to indicate the oracle plan, as well as a random distractor plan for unlikelihood training. During inference, the PGA generates a single summary for each unique content plan returned by the top $K$ beams of the EDU generator. }
\label{fig:system-diagram}
\vskip -0.1in
\end{figure*}


Looking at Figures \ref{fig:salience} and \ref{fig:uniqueness} together, a tradeoff is easily visible. DBS has the most pronounced decline in \textbf{Salience} yet most closely satisfies the \textbf{Uniqueness} property (closest to \textbf{\textcolor{Green}{\texttt{Idealized}}}). We hypothesize that an optimal decoding method should achieve a high degree of \textbf{Uniqueness} while exhibiting minimal \textbf{Salience} degradation across beams.

\section{Plan-Guided Abstraction (PGA)}

At a high-level, we ensure\footnote{This presupposes an abstractive LM with perfect plan adherence. We record adherence but do not require perfection.} Uniqueness by conditioning each candidate on its own unique content plan, and minimize quality degradation by only using the top beam from the abstractive decoder. More specifically, we transform a BART LM into a hierarchical encoder, single-decoder model, which learns to copy extractive content plans at the EDU-level (\S \ref{sec:gen-extract}). Another encoder-decoder model (BART for CNN/DM and NYT, PEGASUS for Xsum) learns to generate the reference given special markers to indicate the content plan (\S \ref{sec:gen-abstract}). Figure \ref{fig:system-diagram} depicts the training procedure for Extract Generation (\textbf{Step 1}, \S \ref{sec:gen-extract}) and Plan-Guided Abstraction (\textbf{Step 2}, \S \ref{sec:gen-abstract}), as well as the end-to-end candidate generation method (\textbf{Step 3}).

\subsection{Generating EDU-Level Plans} \label{sec:gen-extract}

\paragraph{tl;dr.} Inspired by the AREDSUM-SEQ model \citep{bi-etal-2021-aredsum}, which itself is based off the hierarchical encoder from BertSumExt \citep{liu-lapata-2019-text}, we adapt a BART conditional language model such that it is able to generate extractive EDU fragments left-to-right, in the order in which they appear. The decoder uses a copy mechanism for EDUs and a special end of extract token. The special token enables EDU extractive plans to have variable length.

\paragraph{Notation.} A document $D$ can be expressed as a list of $K$ non-overlapping EDU segments: $D=\{s_1, s_2, ..., s_K\}$. A content plan $S$ is a subset of the EDUs in the document: $ S \subset D$. Let $S_t^*$ represent an \emph{ordered} partial extract ending in $s_t$. The probability of adding EDU $s_i$ to $S^*_t$ is modeled as:

\vskip -0.1in
\[ \begin{cases} 
      p(s_i | D, S_t^*) & i \in K, i > t \\
      0 & i \in K, i \leq t \\
   \end{cases}
\]

We note that adding EDUs to an extractive plan in the order in which they appear in the document is non-standard. Most extractive models build summaries in a confidence-first fashion, as in \citet{zhou-etal-2018-neural-document}. We experimented with both in-order and confidence-first and found that the former slightly outperformed.

To encode EDUs, we bracket each EDU with start \texttt{<e>} and \texttt{</e>} tokens. We pass the full document: EDU markers and tokens through a pre-trained BART encoder, and extract hidden states for each EDU with mean pooling over each token within the EDU (including the start and stop tokens): $\{h_{s_1}, ..., h_{s_1}\}$. Then, the EDU representations are modeled by a newly initialized EDU-level BART encoder:

\vskip -0.2in
\begin{multline*}
\{h_{s_1}^{'}, ..., h_{s_K}^{'}, h_{eoe}^{'}\} = \\ENC_{sent}(\{h_{s_1}, ..., h_{s_K}, E(eoe)\})
\end{multline*}

\noindent $E(eoe)$ represents a learned embedding for the end of extract token. Positional embeddings are added to each EDU representation ($h_{s_i}$) to indicate its position in the document, before being passed through the stacked transformer layers in the encoder. At decoder timestep $k$ with hidden state $h_k^{*}$ and partial extract $S_t^*$, each valid next output ($s_i \in S, i > t$ and $eoe$) is scored by a single layer MLP, which can be represented as\footnote{Based on \citet{bi-etal-2021-aredsum}, we experimented with redundancy features, yet it did not improve downstream abstract performance.}:

\vskip -0.1in
\[ \begin{cases} 
    W_o([h_i^{'}; h_k^*]) + b_o & s_i \in S, i > t \\
    W_o([h_{eoe}^{'}; h_k^*]) + b_o & eoe \\
   \end{cases}
\]

\paragraph{Plan Objective.} Given the above probability distribution, we treat the plan generator as a standard LM and train it with maximum likelihood estimation (MLE) of the oracle plan given the source document.

\paragraph{Oracle Labels.} As discussed in \S \ref{sec:motivation}, We use the greedy search algorithm proposed by \citet{nallapati2017summarunner} to generate oracle EDU extractive plans.

\paragraph{Inference.} As a functional LM, we generate distinct EDU extractive plans with beam search.

\subsection{Learning to Abstract from EDU Plans} \label{sec:gen-abstract}

\paragraph{tl;dr.} We fine-tune a separate token-level LM, which learns to generate the reference given an oracle plan, while discouraging it from generating the same reference given a random plan. An MLE loss is added as regularization. During inference, the model receives EDU plans from \S \ref{sec:gen-extract} and generates one abstract per plan with standard beam search.

\paragraph{Decorating inputs.} We implement a simple parameter-efficient method for incorporating an extractive plan. We simply demarcate the EDUs in the plan with special start and end tokens \texttt{<e>} and \texttt{</e>}, whose embeddings are learned during fine-tuning. This is similar yet different from the extractive plan generator. When learning to generate plans, all EDUs are tagged, yet when generating the abstract, only the in-plan EDUs are tagged. Decorating the input is a more flexible approach to incorporating extractive guidance than modifying encoder-decoder attention \citep{saito2020abstractive} and is more parameter-efficient than separately modeling the set of extracted text units \citep{dou-etal-2021-gsum}.

\paragraph{Guided-Abstraction Objective.} We use a likelihood objective for plan-guided abstraction, and to improve plan adherence, add an unlikelihood term \citep{welleck2019neural}, which discourages the model from generating the reference given a random plan:

\vskip -0.2in
\begin{multline} \label{eq:loss}
     \mathcal{L_{GA}} = \lambda log(p(R|D,S_{oracle})) \\+ \lambda log(1 - p(R|D,S_{random}))) \\ + \beta log(p(R|D))
\end{multline}

\noindent $S_{oracle}$ represents the oracle plan for the reference $R$ and $S_{random}$ is a randomly sampled plan of the same length from the set of non-oracle source EDUs. The first two terms encourage the model to rely on the plan when generating an abstract, while the final term is the standard MLE objective (without plan) and acts as a regularization term. $\lambda$ and $\beta$ are scalars controlling the relative weight of the plan adherence versus regularization components on the $\mathcal{L_{GA}}$ loss.

\paragraph{Inference.} The guided-abstractor is trained on oracle extractive plans yet, at inference time, realizes extractive content plans produced by the extract generator from \S \ref{sec:gen-extract}. Standard beam search is used to decode a single abstract for each unique plan.

\section{Experimental Setup}

\paragraph{Datasets.} We use the same datasets as in BRIO \citet{liu-etal-2022-brio}, which are CNN / Dailymail \citep{HermannKGEKSB15, see-etal-2017-get}, the New York Times annotated corpus \citep{sandhaus2008new}, and Xsum \citep{narayan-etal-2018-dont}. The first two are more extractive while Xsum is more abstractive and contains highly noisy references \citep{nan-etal-2021-improving}. We use code from \citet{kedzie-etal-2018-content} for data pre-processing and splitting of the corpus, and treat the archival abstract as the ground-truth reference.

\paragraph{Metrics.} We compare summaries to references with ROUGE 1/2/L F1 \citep{lin2004rouge} and BERTScore F1 \citep{bert-score}. We use the standard PERL ROUGE script for ROUGE scoring with PTB tokenization and lowercasing, as in \citet{liu-etal-2022-brio}. For BERTScore, we use the default model (\texttt{roberta-large}) and settings from the widely-used \texttt{bert-score} Python package\footnote{\textit{roberta-large\_L17\_no-idf\_version=0.3.12(hug\_trans=4.6.1)}}.


\begin{table*}[t]
\setlength{\tabcolsep}{5pt}
\centering
\small
\begin{tabular}{l|cccc|cccc|cccc}
\multirow{2}{*}{\textbf{\texttt{{\makecell{Candidate \\ Method}}}}} & \multicolumn{4}{c}{\textbf{\texttt{CNN/DM}}} & \multicolumn{4}{c}{\textbf{\texttt{NYT}}} &  \multicolumn{4}{c}{\texttt{\textbf{Xsum}}} \\
& \texttt{\textbf{R1}} & \texttt{\textbf{R2}} & \texttt{\textbf{RL}} & \texttt{\textbf{BS}} & \texttt{\textbf{R1}} & \texttt{\textbf{R2}} & \texttt{\textbf{RL}} & \texttt{\textbf{BS}} & \texttt{\textbf{R1}} & \texttt{\textbf{R2}} & \texttt{\textbf{RL}} & \texttt{\textbf{BS}} \\ \hline
\texttt{Top Beam\textsuperscript{$\dag$}} & 44.0 & 21.03 & 37.42 & 86.38 & 54.02 & 35.10 & 50.84 & 89.05 & 47.23 & 24.60 & 39.37 & 91.32 \\ \hline
\texttt{SimCLS\textsuperscript{$\ast$}} & 46.67 & 22.15 & 43.54 &- &- &- &- &- &47.61 &24.57 &39.44 &- \\
\texttt{SummaReRanker\textsuperscript{$\ast$}} & 47.16 &22.55 &43.87 &- &- &- &- &- &48.12 &24.95 &40.00 &- \\
\texttt{BRIO-Ctr\textsuperscript{$\ast$}} & 47.28 &22.93 &44.15 &- &55.98 &36.54  &52.51 &- &48.13 &25.13 &39.80 &- \\
\texttt{SummaFusion\textsuperscript{$\ast$}} & - & - &- &- &- &- &- &- & 47.08 & 24.05 &38.82 &- \\ \hline
\texttt{Beam Search\textsuperscript{$\dag$}} & 45.26 & 22.04 &41.87 &88.52 &55.24 &36.61 & 51.99 & 89.52 & 48.40 & 25.50 & \textbf{40.36} & \textbf{91.46} \\
\texttt{Diverse Beam\textsuperscript{$\dag$}} & 46.98 & 22.90 & 43.85 & 88.95 &54.89 & 36.05 & 51.62 & 89.56 & 47.86 & 24.84 & 39.81 & 91.41 \\
\texttt{Nucleus\textsuperscript{$\dag$}} & 46.57 &23.06 &43.37 & 88.84 & 55.15 & 36.38 & 51.83 & 89.33 & 46.78 & 23.74 & 38.86 & 91.20 \\ \hline \hline
\texttt{\textbf{PGA (ours)}} & \textbf{47.59}\textsuperscript{$\ddag$} & \textbf{23.81}\textsuperscript{$\ddag$} & \textbf{44.33}\textsuperscript{$\ddag$} & \textbf{89.02} & \textbf{57.19}\textsuperscript{$\ddag$} & \textbf{38.55}\textsuperscript{$\ddag$} & \textbf{54.12}\textsuperscript{$\ddag$} & \textbf{89.96} & \textbf{48.44} & \textbf{25.51} & 40.34 & 91.45 \\
\end{tabular}
\vskip -0.1in
\caption{ROUGE-F1, BERTScore (BS) metrics for top-ranked summaries across three datasets. \textbf{Best} results across all rows are \textbf{bolded} and $\ddag$ are statistically significant ($p < .05$) with respect to our internal baselines $\dag$ (Confidence testing is only done for ROUGE scores, not BS). \texttt{Top Beam} represents the conventional single candidate setup, $\ast$: reported results in re-ranking papers. $\dag$: candidates generated by us and re-ranked by available BRIO re-rankers \citep{liu-etal-2022-brio}). Candidates from our PGA method are re-ranked by the same BRIO models to allow for direct comparison with our baselines ($\dag$). } \label{tab:results}
\end{table*}

\paragraph{Baselines.} We generate 16 candidates with different decoding methods: beam search, diverse beam search, and nucleus sampling. We use \texttt{google/pegasus-xsum} for Xsum, \texttt{facebook/bart-large-cnn} for CNN, and fine-tune a BART-Large model on the NYT corpus. For \texttt{NYT}, we fine-tune using a standard MLE loss for up to 10 epochs, choosing the best model based on validation ROUGE score. These are also the checkpoints used to initialize our plan extractor token-level encoder and guided abstractor. We also compare our method to previous work on summary re-ranking. \textbf{SimCLS} \citep{liu-liu-2021-simcls} and \textbf{BRIO-Ctr} \citep{liu-etal-2022-brio} both generate 16 candidates via diverse beam search using the same pre-trained weights as in our work\footnote{Given that we use the same re-ranker and evaluation script, our diverse beam search baseline aims to replicate \textbf{Brio-CTR}.}. The major difference between the papers is that a RoBERTa \citep{liu2019roberta} classifier is used for re-ranking SimCLS, while in BRIO, the model likelihoods are calibrated to ROUGE rankings. \textbf{SummaReranker} \citep{ravaut-etal-2022-summareranker} trains a RoBERTa-based mixture of experts classifier on up to 60 candidates ensembled from multiple decoding methods (beam search, diverse beam search, nucleus sampling, and top-k sampling). We report their best ensemble configuration for CNN and NYT, which uses dataset-specific fine-tuned PEGASUS \citep{zhang2020pegasus} checkpoints from the HuggingFace Transformers library \citep{wolf2020transformers}. \textbf{SummaFusion} \citep{ravaut2022towards} fuses candidate summaries into a single summary. Candidates are generated with diverse beam search from the same PEGASUS checkpoint for Xsum (\texttt{google/pegasus-xsum}).

\paragraph{Training Details.} For the EDU plan generator, we initialize the token-level encoder from fine-tuned summarization checkpoints for each dataset (listed above in \textit{Baselines} paragraph). The EDU-level BART encoder and decoder are randomly initialized to have two layers (using a BART-Large configuration to determine parameter dimensions). For both EDU-Extract and Guided abstract training, we fine-tune with Pytorch Lightning \citep{falcon2019pytorch} for a maximum of 150,000 steps with 200 warmup steps, a learning rate of 1e-5, batch size of 16, and weight decay of $5e-5$. For Xsum, we fine-tune plan-guided abstraction from \texttt{google/pegasus-xsum} and use a learning rate of $1e-4$ and a batch size of $64$. 

For the EDU generator, we select the checkpoint that maximizes the ROUGE score on the validation set. For the Plan-Guided Abstractor, we select the checkpoint that maximizes the oracle-guided abstract ROUGE score. We grid-searched $\lambda$ and $\beta$ from Equation \ref{eq:loss} over $[0, 0.1, 1, 10]$  and selected based on top-ranked validation set summaries. For NYT, we set $\lambda = 1$ and $\beta = 0$ from Equation \ref{eq:loss}. No regularization is needed. For CNN and Xsum, we use more regularization: $\lambda = 1$ and $\beta = 10$. For Xsum, we enforce the last plan beam to be the null-plan (no EDU guidance)\footnote{Given regularization ($\beta > 0$), the model retains its ability to generate without extractive guidance (\texttt{<e>}, \texttt{</e>}) decorators.}.

\paragraph{Decoding Parameters.} For EDU plan generation, we set the min-max plan lengths to 2-20 and use a length penalty of $1.0$ for CNN and NYT, while $2.0$ for Xsum. For plan-guided abstraction, we set a beam size of $4$ for CNN and NYT, while $8$ for Xsum. The baselines and plan-guided models use the same min-max summary lengths and length penalties: 56-142 and 2.0 for CNN, 56-256 and $2.0$ for NYT, and 11-62 and $0.6$ for Xsum. For nucleus sampling, we set $p=0.92$. For diverse beam search, we set the diversity penalty to $1$ and set the number of beams and beam groups equal to the number of candidates (16), as in \citet{liu-etal-2022-brio}.

\paragraph{Re-Rankers.} We obtain top ranked summaries from pre-trained re-rankers supplied from BRIO \citep{liu-etal-2022-brio}. Their CTR model coordinates likelihoods with ROUGE-defined rankings by optimizing the following pairwise margin ranking loss:

\vskip -0.1in

\begin{align}
\small
\label{eq:brio}
\begin{split}
max(0, f(D, \hat{y}_j) - f(D, \hat{y}_i) + (j - i) * \lambda) \forall i, j \in |\hat{Y}|, i < j
\end{split}
\end{align}

\noindent where $\hat{Y}=\{\hat{y}_1, ..., \hat{y}_n\}$ represents an ordered list of summaries: $ROUGE(\hat{y}_i, y) \geq ROUGE(\hat{y}_j, y)$, $\forall i, j \in |\hat{Y}|, i < j$. $f$ represents the length normalized log likelihood of generating the summary.  We use BRIO configurations and default hyper-parameters.


\section{Results} \label{sec:results}

Please refer to Appendix \ref{sec:beam-consistency} for an analysis of the beam consistency of PGA candidates versus baselines.

\paragraph{Re-Ranked Performance.} 

Table \ref{tab:results} shows that the top-ranked summaries of PGA candidate sets consistently outperform. Compared to the best internal baseline method (beam search, diverse beam, nucleus sampling), we see ROUGE-2 F1 percentage advantages of $\bm{.75}$ ($23.81$ versus $23.06$), $\bm{1.94}$ ($38.55$ versus $36.61$), and $\bm{.01}$ ($25.51$ versus $25.50$) on CNN/DM, NYT, and Xsum, respectively. Our PGA method also outperforms the best published results for re-ranked summaries. In particular, across datasets, we see ROUGE-2 F1 percentage advantages of $\bm{.88}$ ($23.81$ versus $22.93$), $\bm{2.01}$ ($38.55$ versus $36.54$), and $\bm{.38}$ ($25.51$ versus $25.13$). The performance gains against our internal baselines ($\dag$ in Table) \ref{tab:results} are significant for CNN/DM and NYT ($\bm{p < 0.05}$), but not for Xsum. Extractive planning may be less useful when reference summaries are shorter and noisier. Xsum references have been shown to contain entity-based ``hallucinations''--content that is unsupported by the input document \citep{narayan-etal-2021-planning, nan-etal-2021-entity}.


\begin{table}[h]
\centering
\small
\begin{tabular}{cl|ccc|c}
& \texttt{\textbf{Method}} & \texttt{\textbf{R1}} & \texttt{\textbf{R2}} & \texttt{\textbf{RL}} & \texttt{\textbf{\# CPs}} \\ \hline
\multirow{4}{*}{\textbf{DCP}} & \texttt{BS} & 41.8 & 19.2 & 35.3 & 6.3 \\
& \texttt{DBS} & 41.5 & 18.9 & 34.9 & 12.7 \\
& \texttt{Nucleus} & 42.0 & 19.4 & 35.3 & 9.9 \\
& \textbf{\texttt{PGA (Ours)}} & 43.6 & 20.8 & 36.9 & 13.0  \\ \hline
\textbf{ECP} & \texttt{EDU Plan} & 43.1 & 20.5 & 36.8 & 16 \\\hline
\end{tabular}
\caption{ Analyzing set statistics for Explicit Content Plans (ECP) versus Derived (DCP). We compare the ROUGE scores of plans vis-a-vis reference, as well as the number of unique content plans (ECP or DCP) from sets of 16. Results shown for CNN / Dailymail test set. } \label{tab:plan-performance}
\vskip -0.1in
\end{table}

\paragraph{Analyzing Content Plans.}
We compare the explicit plans from our EDU-plan generator with Derived Content Plans (DCPs) from our baseline decoding methods, as defined in \S \ref{sec:motivation}, to assess whether or not a dedicated content selector is a better content selector than a derived one. Table \ref{tab:plan-performance} reveals that explicit content plans (ECPs) outperform all DCPs ($43.1$ R1 versus $41.8$ / $41.5$ / $42.0$), except when the DCP is derived from an ECP-guided summary ($43.6$ R1). Using simpler terms, a dedicated content selector chooses more relevant content than the content implied by token-level abstractors, and this performance gain is only overturned when generating an abstract conditioned on these high quality content plans.

\begin{table}[ht]
\centering
\small
\begin{tabular}{l|ccc}
\texttt{\textbf{Method}} & \texttt{\textbf{\makecell{DCP \\ Sent}}} & \texttt{\textbf{\makecell{Summary \\ Sents}}} & \texttt{\textbf{\makecell{Fusion \\ Ratio}}} \\ \hline
\texttt{Beam} & 3.22 & 3.17 & 1.03 \\
\texttt{Diverse Beam} & 3.85 &3.86 &1.02 \\
\texttt{Nucleus} & 3.75 &3.69 &1.03 \\
\textbf{\texttt{PGA (ours)}} & 3.81 &3.69 & 1.05 \\ \hline
\texttt{Reference} & 4.25 &3.76 &1.17 \\
\end{tabular}
\caption{ Fusion ratios: \# of unique source sentences which contain the EDUs in the implied plan (\# DCP Sent), divided by the number of sentences in the summary. } \label{tab:fusion}
\vskip -0.1in
\end{table}

\paragraph{Fusion Analysis.} One of the potential benefits to EDU-based content planning is fusion. Prior work has argued that fusion is desirable for its impact on conciseness, while noting that existing models perform very little fusion \citep{lebanoff-etal-2020-learning}. We measure fusion at the candidate level across decoding methods (including PGA), as well as the summary references, by computing the EDU-level Derived Content Plan (DCP) for each summary, and then recording how many unique source sentences contain the EDUs in this implied plan. To normalize, we then divide it by the number of predicted summary sentences to provide an approximate \texttt{fusion ratio}. Table \ref{tab:fusion} shows that, while PGA has a higher fusion ratio on average than the baselines ($1.05$ versus $1.03, 1.02, 1.03$), model-generated summaries fuse content from fewer sources sentences than human-generated summaries (the Reference fusion ratio is the highest at $1.17$).

\begin{table}[ht]
\centering
\small
\begin{tabular}{l|ccccc}
\texttt{\textbf{Method}} & \texttt{\textbf{Q1}} & \texttt{\textbf{Q2}} & \texttt{\textbf{Q3}} & \texttt{\textbf{Q4}} & \texttt{\textbf{Avg}} \\ \hline
\texttt{Beam} & \textit{47.8} & \textit{46.2} & \textit{44.5} & \textit{42.6} & \textit{45.3} \\
\texttt{Diverse Beam} & \textit{49.2} & \textit{48.0} & \textit{46.0} & \textit{44.7} & \textit{47.0} \\
\texttt{Nucleus} & \textit{48.7} & \textit{47.5} & \textit{45.7} & \textit{44.3} & \textit{46.6} \\ \hline
\texttt{\textbf{Baseline Avg}} & 48.6 & 47.2 & 45.5 & 43.9 & 46.3 \\
\textbf{\texttt{PGA (ours)}} & \textbf{50.1} & \textbf{48.5} & \textbf{46.5} & \textbf{45.3} & \textbf{47.6} \\ \hline \hline
\texttt{\textbf{Avg \% Gain}} & \textbf{\textcolor{Green}{3.09}} & \textbf{\textcolor{Green}{2.75}} & \textbf{\textcolor{Green}{2.20}} & \textbf{\textcolor{Green}{3.19}} & \textbf{\textcolor{Green}{2.81}} \\
\end{tabular}
\caption{ ROUGE-1 F1 for top-ranked summaries on the CNN/DM test set binned into quartiles by summary length. } \label{tab:length}
\vskip -0.1in
\end{table}

\paragraph{Impact of Length.} Previous work has shown that content selection is more difficult as inputs scale \citep{ladhak-etal-2020-exploring}. This would suggest that our approach, which relies on explicit content plans, might scale well to long inputs. To get a sense of the relative impact of the PGA method by length, we bin the CNN test set into quartiles based on the number of EDUs in the source document. In Table \ref{tab:length}, we report average ROUGE-1 F1 scores of top-ranked summaries for the baseline methods and PGA, as well as an average of the baselines (Baseline Avg). The final row (Avg \% Gain) shows the percentage gain for each quartile of moving from Baseline Avg to PGA. The gain is the largest for the fourth quartile ($3.19\%$), yet the increase is not monotonic. The second largest benefit comes from the shortest quartile $3.09\%$. While not conclusive, this analysis suggests that our PGA method could benefit even further from application to long-document and/or multi-document corpora, on which re-ranking methods are largely untested.

\begin{table}[ht]
\setlength{\tabcolsep}{3pt}
\centering
\small
\begin{tabular}{l|ccc|ccc}
& \multicolumn{3}{c}{\textbf{Top Ranked}} & \multicolumn{3}{c}{\textbf{Plan Adherence}} \\
\texttt{\textbf{Method}} & \texttt{\textbf{R1}} & \texttt{\textbf{R2}} & \texttt{\textbf{RL}} & \texttt{\textbf{R}} & \texttt{\textbf{P}} & \texttt{\textbf{F1}} \\ \hline
\texttt{\textbf{PGA (ours) }} & 47.59 & 23.81 & 44.33 & 87.1 & 78.6 & 81.5 \\
\texttt{\textbf{\textcolor{red}{w/o Unlike}}} & 47.43 & 23.48 & 44.16 & 87.2 & 76.5 & 80.3 \\
\end{tabular}
\caption{ Impact of removing the unlikelihood objective from Equation \ref{eq:loss} on the top-ranked summary ROUGE scores and on average adherence to the content plan. } \label{tab:ablation}
\vskip -0.1in
\end{table}

\paragraph{Plan Adherence.} Adherence to the plan is critical to the diversity of PGA outputs given that each candidate is produced from the top beam of the abstractor. If it ignores the provided content plan, all the candidates will be the same. We measure plan adherence by comparing the overlap of DCPs (the implied plan \emph{realized} by the abstractor)  versus ECPs (the plan \emph{provided to} the abstractor).  In particular, we measure the recall, precision, and F1-overlap metrics.  Additionally, we train a PGA model without the unlikelihood objective in Equation \ref{eq:loss} to determine its importance to plan adherence and the ROUGE scores of downstream re-ranked candidates. Table \ref{tab:ablation} shows the ablated model's performance vis-a-vis the PGA model trained with the unlikelihood loss. The top ranked ROUGE-1 is hurt by removing the loss ($47.59$ versus $47.43$ R1), and the abstractor also adheres less to the ECP ($81.5$ versus $80.3$). While the differences are minor, control could be important for human-in-the-loop use cases, in which a user highlights an extractive plan and expects a summary which focuses on these highlights.

\begin{table}[h]
\setlength{\tabcolsep}{10pt}
\centering
\small
\begin{tabular}{l|cc}
\texttt{\textbf{Method}} & \texttt{\textbf{ACU}} & \texttt{\textbf{\textit{n}ACU}} \\ \hline
\texttt{BART} \citep{lewis-etal-2020-bart} & 0.3671 & 0.2980 \\
\texttt{BRIO-Mul} \citep{liu-etal-2022-brio} & 0.4290 & 0.3565 \\
\texttt{T0} \citep{sanh2021multitask} & 0.2947 & 0.2520 \\
\texttt{GPT-3} \citep{brown2020language} & 0.2690 & 0.2143 \\ \hline
\texttt{Diverse Beam Search} & 0.3683 & 0.3261 \\ \hline
\texttt{\textbf{PGA (ours)}} & \textbf{0.4421} & \textbf{0.3650} \\
\end{tabular}
\caption{Human evaluation using the ACU protocol \citet{fabbrirose2022}; the first four rows are copied from their Table 7. Diverse Beam represents our best re-ranking baseline according to ROUGE. \textbf{PGA (ours)} represents a state of the art improvement in reference-based human assessment. } \label{tab:human_eval}
\vskip -0.1in
\end{table}

\paragraph{Human Evaluation.}
To verify the ability of our approach to better capture salient information found in reference summaries, we perform a human evaluation study using the Atomic Content Unit (ACU) protocol introduced in \citet{fabbrirose2022}. In this protocol, atomic facts are extracted from reference summaries and matched with system summaries; the average number of matched units constitutes the recall-focused ACU score, and a length normalized ACU score (\textit{n}ACU) is also reported. We apply this protocol on MTurk and filter workers from the US/UK with 98\% HIT approval and provide a pay-rate of \$12/hour. We use the provided reference ACUs from a 100-example subset from \citet{fabbrirose2022} and achieve a Krippendorf alpha of $0.70$ over three annotators. We compare against our \texttt{Diverse Beam Search} baseline in addition to the four systems from the ACU paper: \texttt{BART}, \texttt{BRIO-Mul}, \texttt{T0}, and \texttt{GPT-3}. As shown in Table \ref{tab:human_eval}, \texttt{PGA} top-ranked summaries outperform summaries from the state of the art supervised\footnote{While included, it is not fair to compare \texttt{PGA} to zero-shot results from \texttt{GPT-3} or \texttt{T0}. The ACU evaluation framework is reference-based, which \emph{strongly} favors supervised models.} model (\texttt{BRIO-Mul}) with respect to un-normalized and length-normalized (ACU / \emph{n}ACU) matching of ACUs between reference and system summaries: $0.4421$ / $0.3650$ for \texttt{PGA} versus $0.4290$ / $0.3565$ for \texttt{BRIO-Mul}.

\section{Guiding GPT with EDU Plans} \label{sec:gpt}

\paragraph{Background.} To date, GPT models \citep{brown2020language, ouyang2022training} have only been evaluated as summarizers in the conventional single candidate setup \citep{zhang2023benchmarking}. In zero and few-shot settings, GPT summaries have been shown to under-perform fine-tuned models with regards to reference-based metrics, yet over-perform according to human judgments \citep{goyal2022news, fabbrirose2022}.

\paragraph{Diverse Prompt-Then-Rank as Alternative to ICL.} To better align closed-source LLMs, such as GPT, to labeled data, in-context learning (ICL) \citet{brown2020language, min-etal-2022-rethinking} has been shown to help. Yet, closed source LLMs can also be adapted to a task by eliciting diverse outputs and then applying a task-specific, smaller re-ranker (e.g., BRIO). ICL and diverse prompt-then-rank can be complementary.

\paragraph{Experimental Setup.} We sample a set of 1,000 summaries at random from the CNN/DailyMail test set and prompt GPT-3.5 \citep{ouyang2022training} to generate summaries. Similarly to \textbf{Top Beam} in Table \ref{tab:results}, we include a single candidate baseline (\texttt{Single}) with the instruction from \citet{goyal2022news, zhang2023benchmarking}: \texttt{Summarize the article in three sentences.} For re-ranking baselines, we generate $16$ diverse candidates by separately increasing the temperature $0.3 \rightarrow 0.7$ (\texttt{Temperature Sampling}), and sampling from a $0.8$ nucleus (\texttt{Nucleus Sampling}). To implement PGA, we decorate the source article with EDU tags \texttt{<e> ... </e>} and instruct GPT to summarize only the text within the tags. Specifically, we instruct it to \texttt{Summarize the content in between the HTML tags <e> and </e> in one to three sentences.} As with \texttt{Single}, we set the temperature to $0.3$. In all cases, we randomly sample 3 examples from the training set to be used as in-context exemplars. We compute a different random sample for each test case to encourage diversity, as in \citet{adams2023desired}. For PGA ICL, we decorate articles with the oracle plan.

\begin{table}[h]
\centering
\small
\begin{tabular}{l|ccc}
\textbf{\texttt{Candidate Method}} & \textbf{\texttt{R1}} & \textbf{\texttt{R2}} & \textbf{\texttt{RL}} \\ \hline
\texttt{Single} & 40.84 & 17.30 & 37.07 \\
\texttt{Temperature Sampling} & 42.51 & 19.17 & 38.73 \\
\texttt{Nucleus Sampling} & 42.43 & 19.06 & 38.65\\ \hline
\textbf{\texttt{PGA (ours)}} & \textbf{43.56} & \textbf{20.11} & \textbf{39.95}  \\
\end{tabular}
\caption{ROUGE-F1 metrics for top-ranked GPT-3.5 summaries on a random 1k subset of the CNN/DailyMail test set. \texttt{ Single} represents a single candidate baseline (similarly to \texttt{Top Beam} in Table \ref{tab:results}). The others produce 16 candidates, which are then re-ranked with BRIO. } \label{tab:gpt-results}
\vskip -0.1in
\end{table}

\paragraph{Results.} As shown in Table \ref{tab:gpt-results}, \textbf{PGA} outperforms all single and diverse candidate methods: $43.56$ ROUGE-1 F1 versus $40.84 / 42.51 / 42.43 $ for the baselines. Please refer to Appendix \ref{app:prompt} for a depiction of the prompt and sample plan-guided output. We publicly release all GPT-3.5 candidates to support RLHF \citep{stiennon2020learning} or calibration \citep{zhao2023slic}\footnote{Available for download on the HuggingFace Datasets Hub under the name: \href{https://huggingface.co/datasets/griffin/cnn-diverse-gpt-3.5-summaries}{griffin/cnn-diverse-gpt-3.5-summaries}.}.

\section{Conclusion} \label{sec:conclusion}
In this paper, we demonstrate that offloading content selection to a dedicated extractor, rather than relying on the decoder to perform both content selection and surface realization, can lead to better \emph{and} more diverse content selection across beams, which ultimately leads to increased ROUGE scores for top-ranked summaries after applying a re-ranker. EDU plan-guided abstraction exhibits other encouraging traits, such as an increased level of fusion and scalability to longer inputs.

\begin{figure*}[t]
\centering
\includegraphics[width=\linewidth]{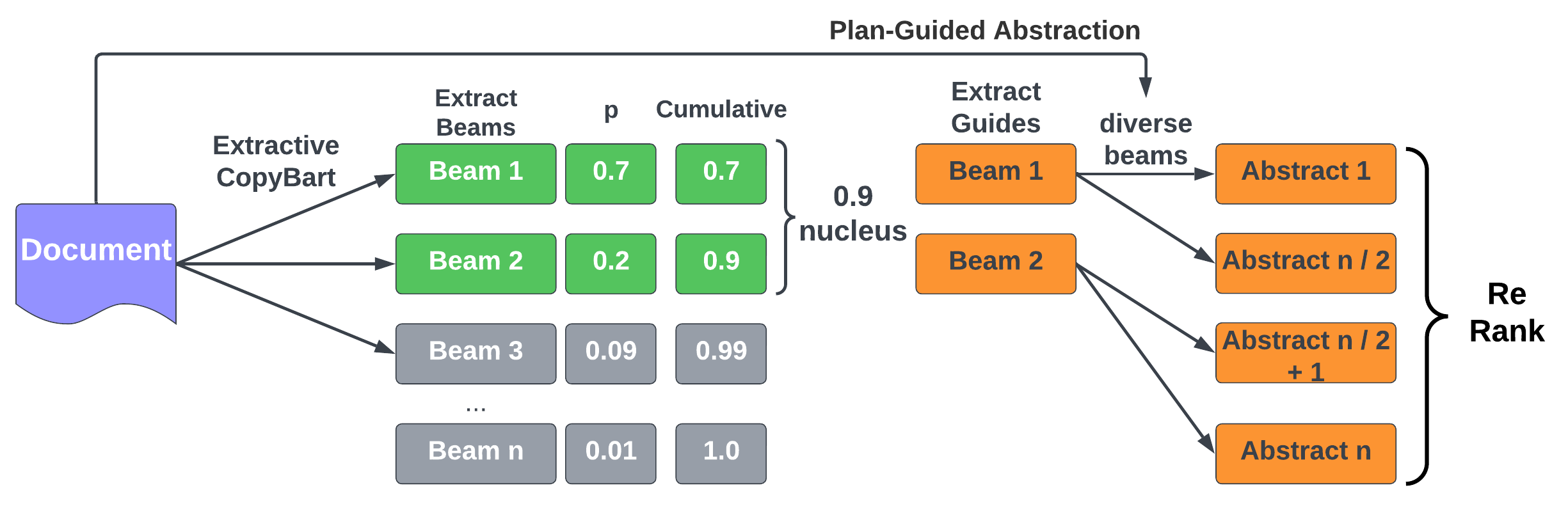}
\caption{Future work could involve generating plan-guided abstracts from a dynamic nucleus of extracts. }
\label{fig:nucleus}
\end{figure*}

\section{Limitations} \label{sec:limitations}

Our findings are primarily based on ROUGE score, which is a noisy, unstable metric with well-studied limitations \citep{schluter-2017-limits}. To address this, however, we conduct a human evaluation to support our findings. In both automatic and human annotation settings, we base our evaluations on naturally occurring references, which have been shown to be silver-standard \citep{gehrmann2022repairing, wan2022factpegasus, adams-etal-2022-learning}. We hope that our work on PGA--a method to generate high-quality diverse candidates--can be applied to new domains (e.g., \citep{gliwa-etal-2019-samsum,adams-etal-2021-whats,deyoung-etal-2021-ms}) and reference-free learning objectives (e.g., RLHF and calibration). Also, our candidate generation method requires two models, which is less elegant and computationally efficient than an end to end solution combining planning and surface realization.

Lastly, PGA treats all content plans as equally likely (each plan is given one abstractive beam). Yet, there is an unexplored trade-off between exploration and exploitation. Should higher-confidence content plans receive more candidates? Future work should explore a generating diverse abstracts from a dynamic nucleus of extracts, which would allow for the generation of many abstracts from only a few extracts when confident (e.g. short documents), while exploring more diverse content when the extractive generator is less confident. We sketch out such a potential system in Figure \ref{fig:nucleus} with a made-up nucleus probability of $0.9$.




\bibliography{anthology,custom}
\bibliographystyle{acl_natbib}

\appendix

\section{Beam Consistency} \label{sec:beam-consistency}

\begin{figure}[t]
\centering
\includegraphics[width=\linewidth]{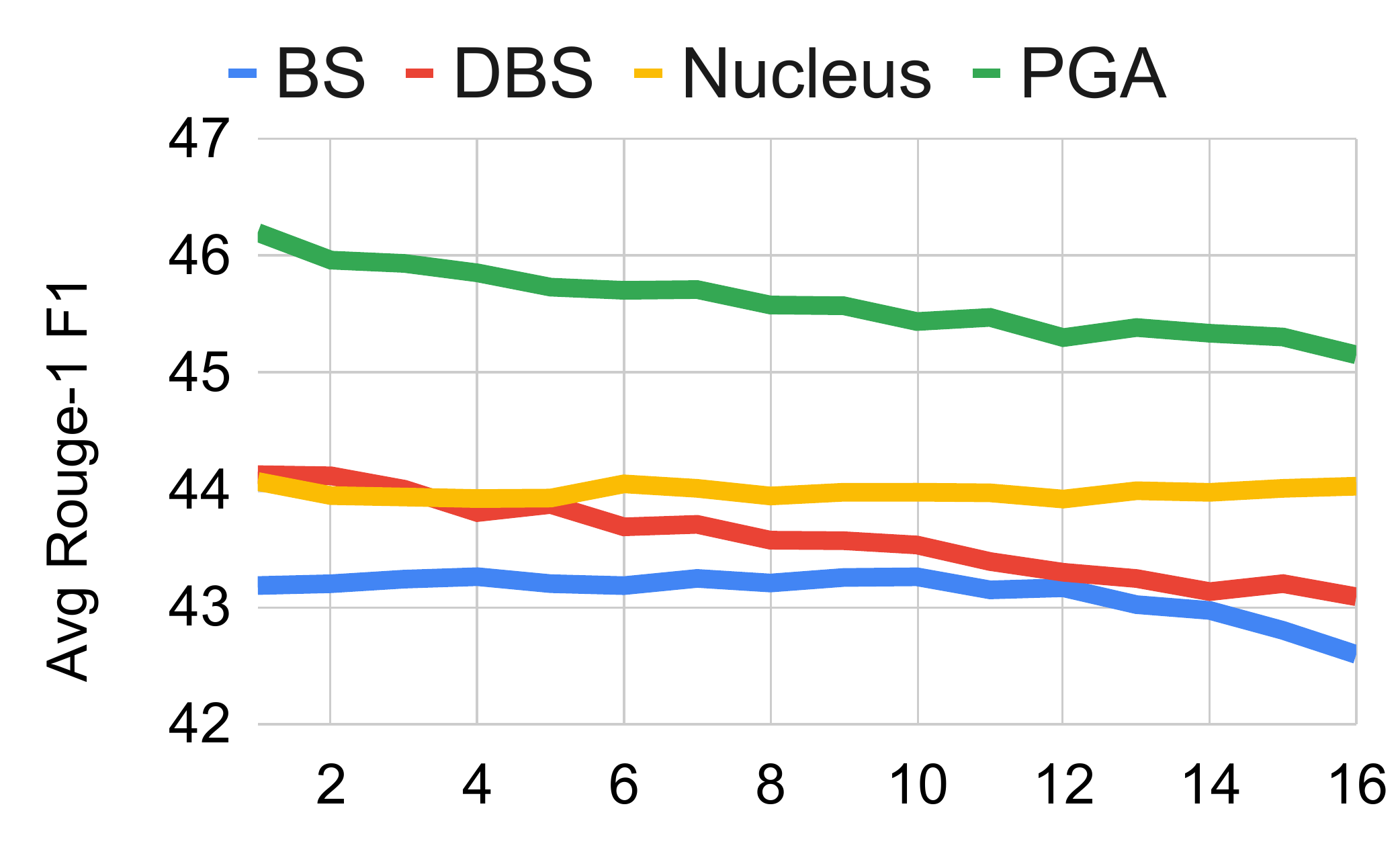}
\caption{ Average ROUGE-1 F1 by beam for the CNN test set. }
\label{fig:rouge-by-beam}
\end{figure}

\begin{figure}[t]
\centering
\includegraphics[width=\linewidth]{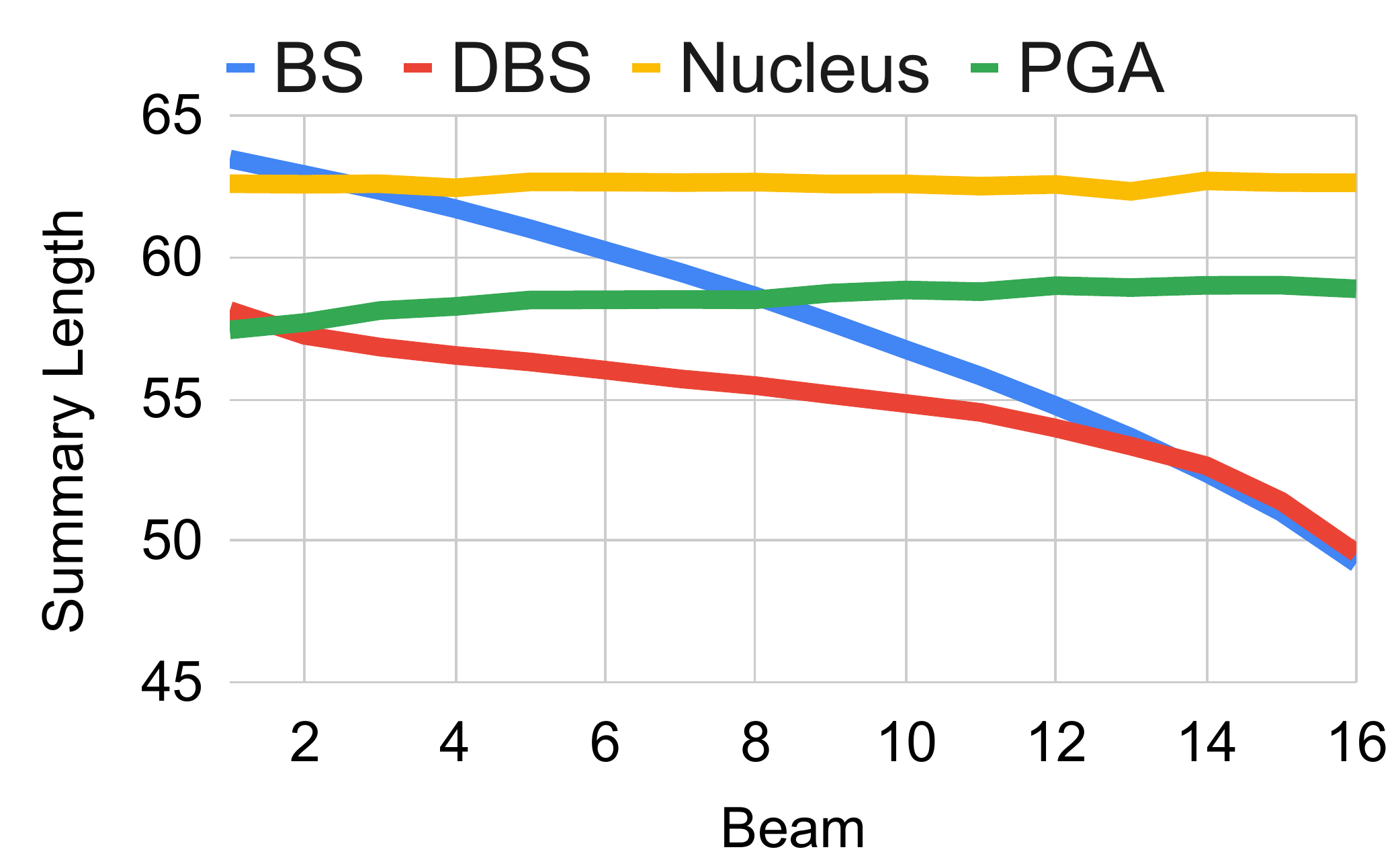}
\caption{ Average length by beam for the CNN test set. }
\label{fig:length}
\end{figure}

\paragraph{Consistency across beams.} A primary benefit to PGA is that each candidate is selected from the top beam. To see whether this leads to more consistency across candidates, we analyze average ROUGE-1 F1 scores by beam, as well as average lengths on the CNN / Dailymail test set. Figure \ref{fig:rouge-by-beam} shows that, on the CNN / Dailymail test set, our PGA candidates obtain higher average ROUGE scores across beams than all other methods. In fact, the last beam PGA has a higher average ROUGE-1 score than the top beam of all baseline methods. Figure \ref{fig:length} shows that nucleus and PGA candidates are more stable length-wise than beam search (regular and diverse). For nucleus, the stability comes from the fact that each candidate is produced by the same sampling procedure. For beam search, the sharp drop-off suggests that length variability may be driving diversity, rather than content selection (as evidenced by DCP redundancy from Table \ref{tab:plan-performance}).

\begin{figure*}[t]
\centering
\includegraphics[width=\linewidth]{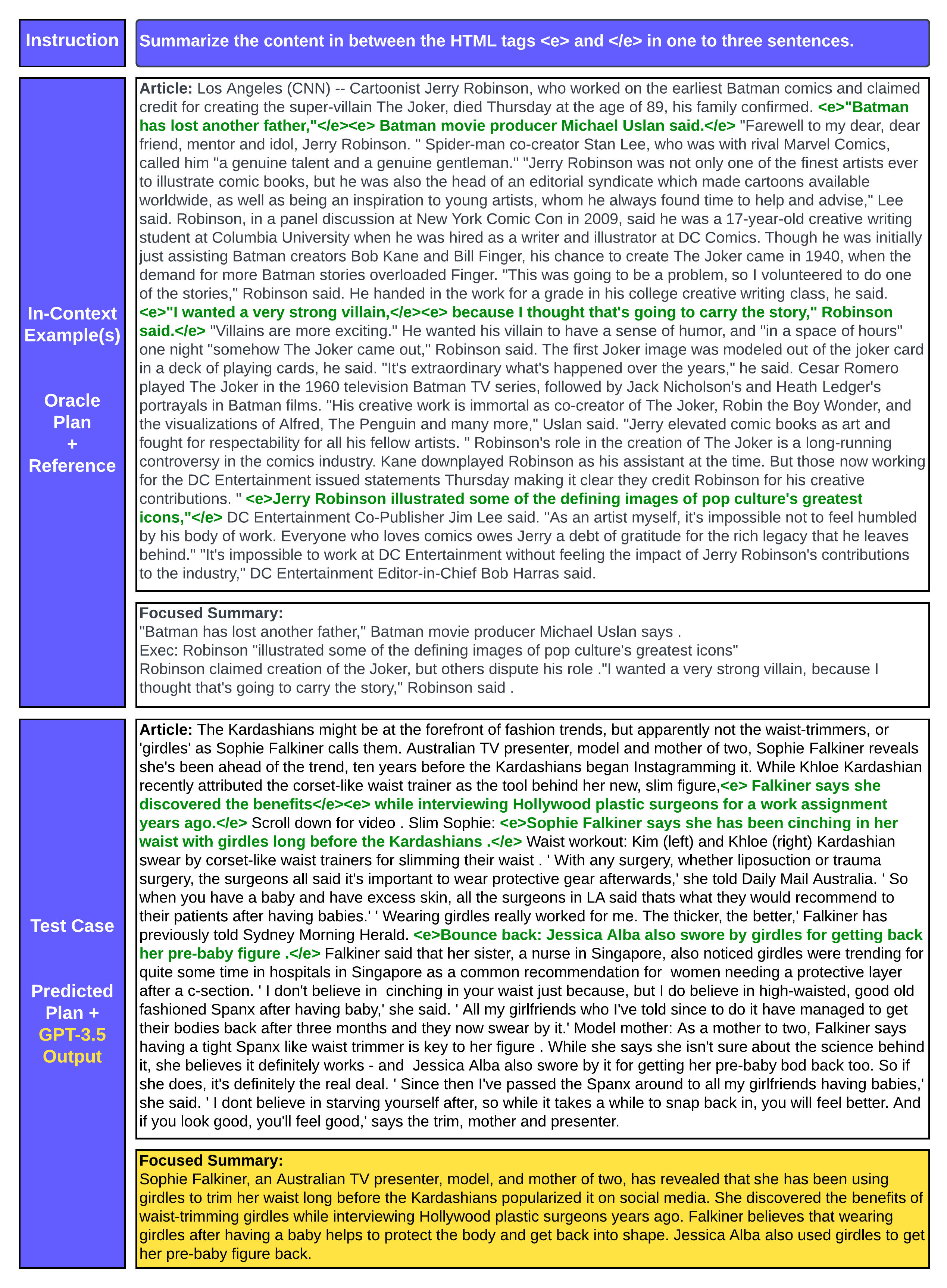}
\caption{ GPT-3.5 Prompt. The instruction is to summarize the content within the \texttt{<e>...</e>} tags. In-Context examples are constructed using oracle EDU plans. Then, GPT-3.5 is given a test case and generates its own \textbf{Focused Summary}, which is highlighted in yellow. GPT-3.5 generates 16  focused summaries based on 16 unique plans. }
\label{fig:gpt}
\end{figure*}

\section{Prompting GPT-3.5 with PGA}
\label{app:prompt}

Figure \ref{fig:gpt} (below) shows the prompt instruction, an in-context example, and an example output from the CNN/DM test set. For the results in \S \ref{tab:gpt-results}, three in-context examples are sampled from the test set.


\end{document}